\documentclass[10pt,twocolumn,letterpaper]{article}

\usepackage{cvpr}              
\usepackage[numbers]{natbib}
\usepackage{multirow}
\definecolor{cvprblue}{rgb}{0.21,0.49,0.74}
\usepackage[pagebackref,breaklinks,colorlinks,allcolors=cvprblue]{hyperref}

\def\paperID{15601} 
\def\confName{CVPR}
\def\confYear{2025}

\title{Gradient-Guided Parameter Mask for Multi-Scenario Image Restoration Under Adverse Weather }

\author{Jilong Guo\\
Harbin Engineering University\\
{\tt\small gjl0503@hrbeu.edu.cn}
\and
Haobo Yang\\
Mohamed bin Zayed University of AI\\
{\tt\small haobo.yang@mbzuai.ac.ae}
\and
Mo Zhou\\
Tsinghua  University\\
{\tt\small zhou-m21@mails.tsinghua.edu.cn}
\and
Xinyu Zhang\\
Tsinghua  University\\
{\tt\small xyzhang@tsinghua.edu.cn}
}

\begin{document}

\maketitle

\begin{abstract}

Removing adverse weather conditions such as rain, raindrop, and snow from images is critical for various real-world applications, including autonomous driving, surveillance, and remote sensing. However, existing multi-task approaches typically rely on augmenting the model with additional parameters to handle multiple scenarios. While this enables the model to address diverse tasks, the introduction of extra parameters significantly complicates its practical deployment. In this paper, we propose a novel Gradient-Guided Parameter Mask for Multi-Scenario Image Restoration under adverse weather, designed to effectively handle image degradation under diverse weather conditions without additional parameters. Our method segments model parameters into common and specific components by evaluating the gradient variation intensity during training for each specific weather condition. This enables the model to precisely and adaptively learn relevant features for each weather scenario, improving both efficiency and effectiveness without compromising on performance. This method constructs specific masks based on gradient fluctuations to isolate parameters influenced by other tasks, ensuring that the model achieves strong performance across all scenarios without adding extra parameters. We demonstrate the state-of-the-art performance of our framework through extensive experiments on multiple benchmark datasets. Specifically, our method achieves PSNR scores of 29.22 on the Raindrop dataset, 30.76 on the Rain dataset, and 29.56 on the Snow100K dataset. Code is available at: \href{https://github.com/AierLab/MultiTask}{https://github.com/AierLab/MultiTask}.

\end{abstract}

\begin{figure}[ht!]
    \centering
    \includegraphics[width=0.5\textwidth,clip,trim=3 3 3 3]{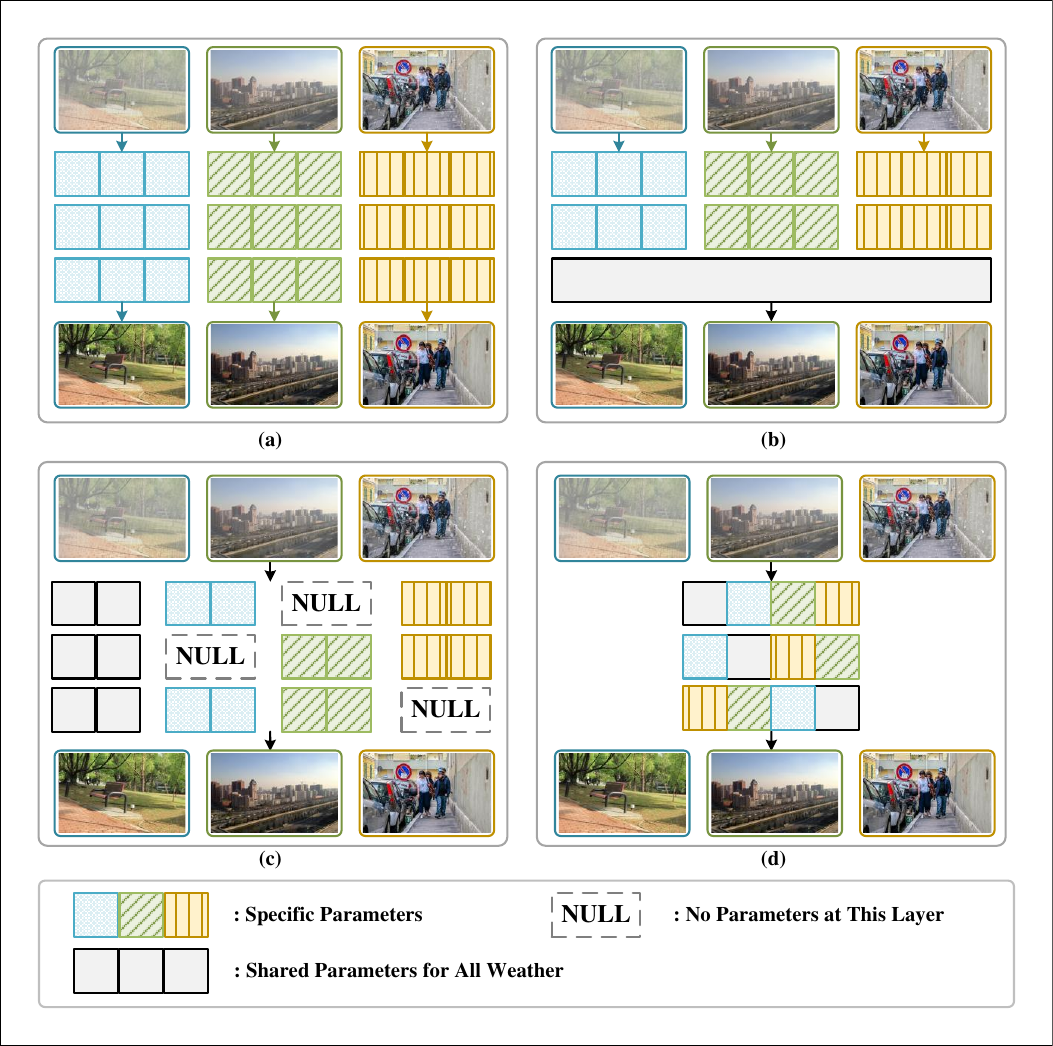} 
    \caption{Illustration of the proposed method and the currently existing solutions. (a) The weather-specific methods; (b) the method
of \cite{li2020all}; (c) methods of \cite{zhu2023learning}; (d) our method,}
    \label{fig:method}
\end{figure}
\section{Introduction}

Adverse weather conditions like rain, snow, and raindrops severely degrade image quality, challenging applications such as autonomous driving, surveillance, and remote sensing\cite{zhang2023perception,vargas2021overview}. These weather-related artifacts can obscure vital visual information, leading to diminished performance in image analysis and interpretation tasks\cite{rothmeier2021let,meister2022designing}. Therefore, developing effective image restoration techniques that can mitigate these adverse effects is essential for enhancing the reliability and effectiveness of systems relying on accurate visual data.

With the advancement and application of deep learning technologies, the widespread use of convolutional neural networks (CNNs)\cite{zeng2020hyperspectral,soh2022variational,su2022survey} and transformers\cite{ali2023vision,zamir2022restormer,liang2021swinir} has led to significant progress in image restoration methods. These sophisticated architectures are particularly adept at capturing intricate features within images, thereby facilitating the effective removal of a multitude of artifacts and degradations. For instance, CNNs\cite{karavarsamis2022survey,rutten2023deep} leverage their powerful local feature extraction capabilities to identify and restore subtle changes resulting from adverse weather conditions. In contrast, transformers\cite{shi2023spatial,choi2023transformer} enhance the modeling of long-range dependencies through global attention mechanisms, making them particularly efficient in addressing large-scale disturbances. The integration of these technologies not only improves the accuracy of image restoration but also significantly boosts processing speed, facilitating real-time applications. Nevertheless, despite the impressive performance of existing methods on individual tasks, most are optimized for single weather conditions, resulting in inefficiencies when confronted with diverse adverse scenarios. 

Recent research \cite{zhu2023learning,li2020all} highlights the general and specific characteristics of image degradation under various weather conditions, which has led to the development of a unified deep model to eliminate weather-related artifacts. However, these solutions have limitations in practical applications. Firstly, as shown in \cref{fig:method}(b), Li et al.\cite{li2020all} enhance the network's recovery capabilities across different scenes by introducing distinct encoders. However, when the dataset distribution across scenes is uneven, smaller datasets can be adversely affected by larger ones, leading to a degradation in performance. Secondly, as illustrated in \cref{fig:method}(c), Zhu et al.\cite{zhu2023learning} achieve efficient image recovery across multiple scenes by adaptively adding parameters during collaborative training on multi-scene data. While this approach improves the model's applicability to various scenarios, the introduction of additional parameters complicates its deployment in real-world applications.

In this work, we propose a Gradient-Guided Parameter Mask for Multi-Scenario Image Restoration Under
Adverse Weather, designed to effectively tackle image degradation caused by various weather conditions, including rain, raindrops, and snow. Our approach introduces a novel mask strategy that partitions model parameters into shared and task-specific components by analyzing the gradient changes induced by the common parameters during training across different weather scenarios. Specifically, we classify parameters into common and task-specific subsets based on the magnitude of the gradient variations induced by the training data for each weather condition. Common parameters capture general features that are consistent across all weather conditions, while task-specific parameters are adapted to the unique characteristics of each specific weather scenario.
This method addresses the challenge of overfitting caused by limited data in individual weather scenarios by leveraging common parameters, which complement the task-specific parameters. The introduction of the mask ensures that the parameter updates for each scenario are isolated from interference caused by other weather conditions, leading to improved image restoration performance. Additionally, by avoiding the addition of extra parameters, our approach ensures that the model remains lightweight, making it highly suitable for deployment in real-time applications, such as autonomous driving, where computational efficiency is crucial.

The key contributions of this work are as follows:
\begin{itemize}

     \item We propose a novel masking method that accomplishes the image restoration tasks without adding extra parameters, ensuring efficient performance.
    \item Our efficient gradient-guided parameter mask effectively decouples task-specific parameters, mitigating interference across diverse weather scenarios and enhancing performance. 
    \item Our approach achieves state-of-the-art performance on multiple datasets. Furthermore, the efficacy of the proposed method is validated on various adverse weather scenarios.
\end{itemize}
\section{Related Work}
    In the field of autonomous driving, image recovery in adverse weather scenes has always been a core issue that has attracted much attention\cite{yurtsever2020survey}.In response to the degradation of image quality caused by different bad weather conditions, previous studies have extensively discussed a variety of image restoration tasks,including deraining\cite{kozar2023recovery,yang2022rain,chen2022lightweightderain,fu2023continual,hu2021single,jiang2020multi,wang2020model,xiao2022image,xiao2021improving,yang2020single},desnowing\cite{chen2020jstasr,zhang2021deep,li2021online,jaw2020desnowgan,zhang2021dual}, and raindrop removal\cite{luo2020weakly,yan2022feature,lin2020x,li2020all}.

\subsection{Rain Removal}

To tackle the challenge of rain streak removal, several approaches leverage Convolutional Neural Networks (CNNs) for effective deraining \cite{ragini2022rain}, \cite{quan2021removing}, utilizing the architecture and module integration of deep neural networks to eliminate rain artifacts. Additionally, Generative Adversarial Networks (GANs) have been employed to efficiently remove raindrops \cite{ding2021rain}, \cite{zhang2019image}. Beyond these, various techniques such as adversarial learning \cite{li2019heavy}, transfer learning \cite{huang2021memory, ye2022unsupervised}, frequency priors \cite{huang2022deep}, and data generation \cite{ni2021controlling} have been explored to enhance model performance.
The Single Image Rain Removal (SIRR) method \cite{lin2020sequential} focuses on learning rain patterns from images through CNNs, enabling effective removal of raindrop effects. CycleGAN-based architectures are also widely utilized for rain removal, where the network learns image transformation via cycle consistency, thereby significantly improving the removal of rain streaks \cite{tang2022multi}. Furthermore, approaches incorporating attention mechanisms \cite{zhang2021dual} have shown to enhance performance by selectively focusing on the rain-affected regions, leading to optimized deraining results.

\subsection{Raindrop Removal}


In recent years, significant progress has been made in the field of raindrop removal in image processing and computer vision. In \cite{9040628}, light-field images and depth maps were used for accurate raindrop detection and subsequent image restoration. \cite{8578361} introduced an iterative contrastive learning framework to improve raindrop removal through incremental optimization. To reduce the dependence on paired training samples, \cite{luo2020weakly} proposed a weakly supervised approach using image-level annotations. In \cite{9577896}, the simultaneous removal of raindrops and rain streaks was demonstrated, improving the restoration quality without adding computational burden. Furthermore, \cite{10378538} presented a sparse sampling transformer with an uncertainty-driven strategy for unified raindrop and rain streak removal, demonstrating the potential of deep learning. In \cite{9707527}, RainGAN was introduced, an unsupervised framework that utilizes decomposition and composition for effective raindrop removal without paired training data. These innovations not only improve the efficiency and accuracy of raindrop removal, but also broaden its applicability in challenging weather conditions.

\subsection{Snow Removal}

Concerning snow removal, various approaches have been proposed to alleviate the impact of snowflakes on visual content. \cite{zhang2021deep} propose a Deep Dense Multi-Scale Network (DDMSNet) for snow removal by exploiting semantic and depth priors. In refinement stage based on generative adversarial networks (GANs) \cite{jaw2020desnowgan} is proposed to further improve the visual quality of the resulting snow-removed images and make a refined image and a clean image indistinguishable by a computer vision algorithm to avoid the potential perturbations of machine interpretation.DCSNet\cite{9897458} is an effective method for capturing the diversity of snowflakes and removing snow layers in stages. This is achieved by adaptively fusing a feature pyramid structure and a progressive restoration module. Furthermore, \cite{10070780} employs a dual gradient strategy to precisely locate snowflakes using gradient activation maps and edge maps. By employing a designed mask estimation network and a transparency-aware context restoration network, it achieves accurate snowflake removal and restoration of image context information.



\subsection{Multi-Task Learning (MTL)}

Recognizing the interconnected nature of weather-related image restoration tasks, researchers have increasingly explored multi-task learning (MTL) paradigms\cite{karavarsamis2022cross,sun2022landslide}. By jointly optimizing models for various tasks, such as rain, raindrop, and snow removal, MTL aims to enhance the overall efficiency and robustness of image restoration algorithms. This collaborative approach facilitates a more unified and effective solution to address the complexities inherent in diverse weather conditions.For example, the AIRFormer model proposed in \cite{10196308} serves as a paradigm for multitask learning. By introducing frequency-guided transformer encoders and decoders, this model achieves integrated processing of image restoration under different weather conditions. The frequency information in the model is used as cross-task shared knowledge, which helps to improve performance across different tasks. Similarly, \cite{10506517} and \cite{Wang2024LRBT} adopt the idea of multitask learning. The former obtains cross-task general feature representations through mask-based pre-training on large datasets, while the latter achieves joint optimization of multiple removal tasks through local reasoning and back-projection mechanisms.

\section{Methodology}

\begin{figure}[!t]
    \centering
    \includegraphics[width=0.5\textwidth,clip,trim=3 3 3 3]{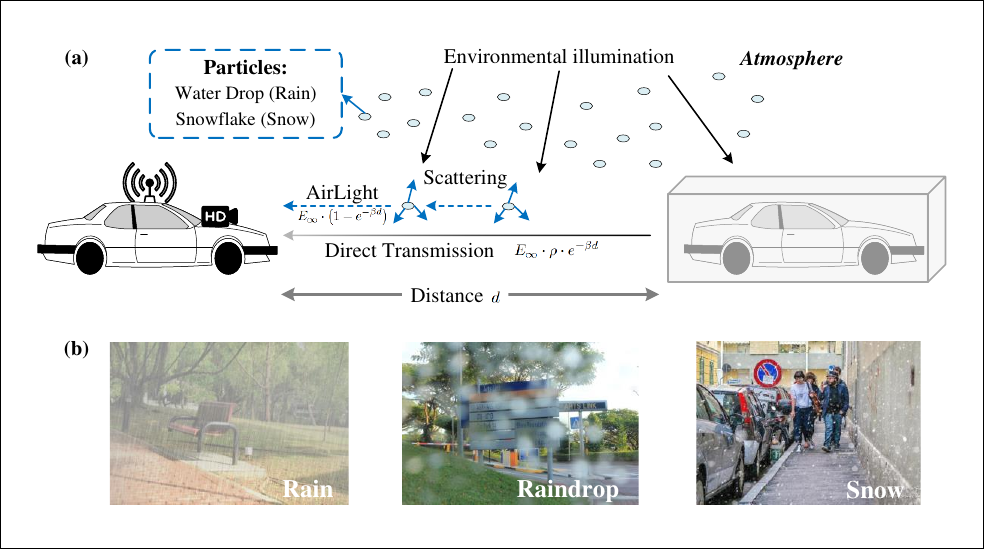} 
    \caption{(a) The environmental illumination of various weather conditions. (b) Real-world scenes of
different weather conditions.}
    \label{fig:atmospheric}
\end{figure}

\begin{figure*}[!ht]
    \centering
    \includegraphics[width=\textwidth,clip,trim=3 3 3 3]{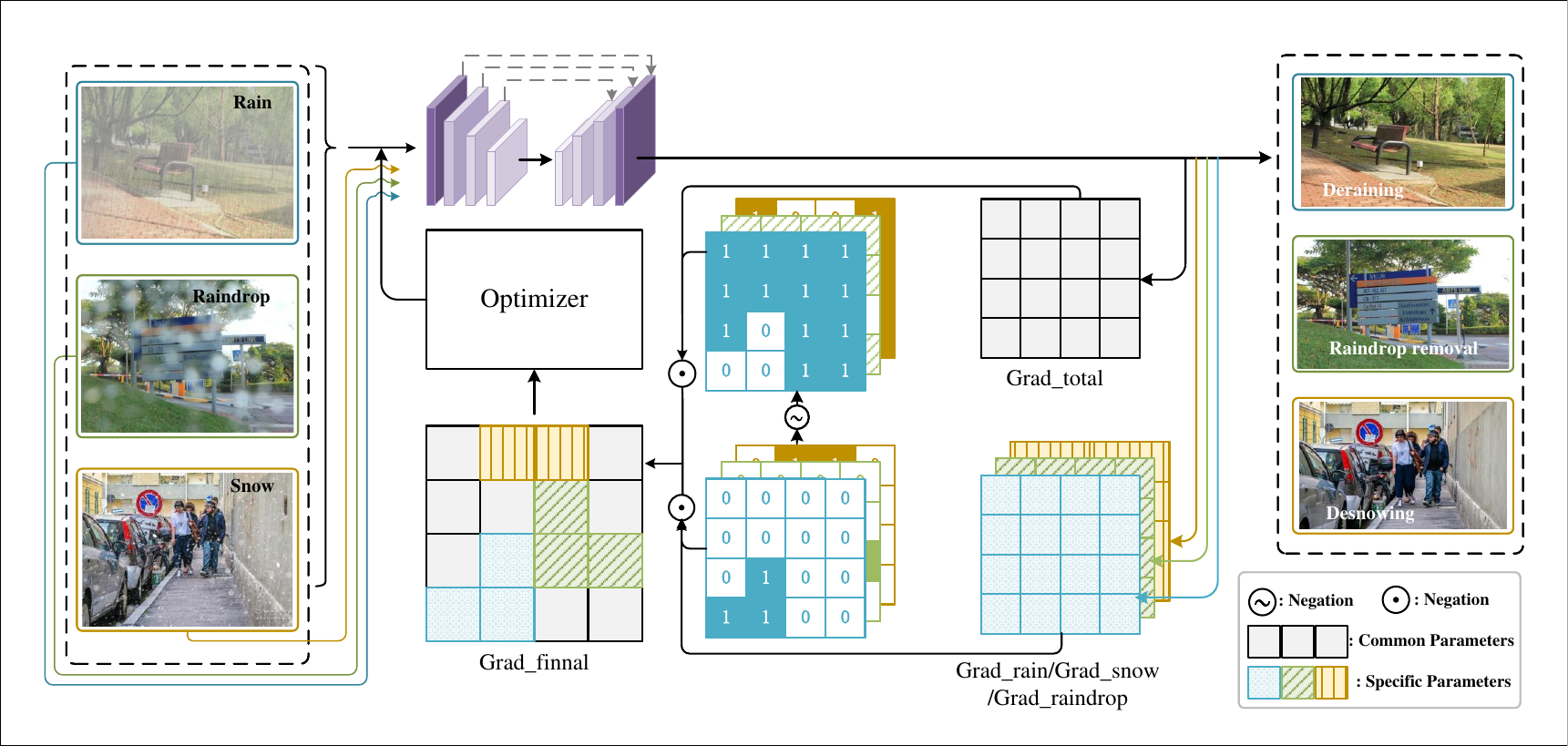} 
    \caption{Illustration of the Gradient-Guided Parameter Mask strategy. The model parameters are partitioned into common and task-specific components: (a) common parameters \( \Theta_c \) retain stable features beneficial across multiple weather scenarios; (b) task-specific parameters \( \Theta_t \) are adaptively learned for individual adverse conditions (e.g., rain, snow, raindrop), enhancing restoration performance while maintaining computational efficiency.}
    \label{fig:adaptive_parameter_masking}
\end{figure*}

To address the challenges posed by image degradation in diverse adverse weather conditions, we propose an \textbf{Gradient-Guided Parameter Masking (APM)} framework for our multi-Scenario adverse weather restoration network. This network is designed to efficiently restore images degraded by weather artifacts such as rain, snow, and raindrop, using a unique parameter masking approach. Our method introduces three distinct masks corresponding to each adverse weather condition, enabling the model to adaptively leverage general and scenario-specific features without increasing the total parameter count, thereby maintaining computational efficiency suitable for real-time applications.

\subsection{Image Degradation Under Adverse Weather}

Adverse weather conditions, including rain, snow, and raindrop, significantly degrade the quality of captured images \cite{narasimhan2003contrast}. This degradation arises from the complex interactions between light and atmospheric particles, which alter the transmission and scattering of light as it travels from the scene to the observer \cite{an2024comprehensive}. The atmospheric scattering model \cite{nayar1999vision}(in \cref{fig:atmospheric}) describes the imaging formulation under such conditions, indicating that the irradiance received from a particular scene point by the photographic sensor is the sum of direct transmission (attenuation) and scattered light:

\begin{equation}
E = E_{\infty} \cdot \rho \cdot e^{-\beta d} + E_{\infty} \cdot \left(1 - e^{-\beta d}\right),
\end{equation}

where \( E \) represents the total irradiance received at a specific point in the scene, \( E_{\infty} \) denotes the intensity of light from the sky, \( \rho \) is the normalized radiance of the scene point, \( \beta \) is the scattering coefficient, and \( d \) represents the optical distance from the scene point to the observer. The direct transmission term \( E_{\infty} \cdot \rho \cdot e^{-\beta d} \) quantifies the attenuation of light, while the airlight term \( E_{\infty} \cdot \left(1 - e^{-\beta d}\right) \) encapsulates the effects of scattering on image quality.

Despite the distinct degradation characteristics associated with each weather phenomenon, they exhibit commonalities attributable to the fundamental physical processes governing light behavior. Specifically, all these weather conditions involve the dual processes of light attenuation and scattering, which manifest similarly within the framework of the atmospheric scattering model. This insight allows us to leverage the common features present across multiple weather scenarios to optimize and adjust model parameters.

\subsection{Gradient-Guided Parameter Mask}

To effectively harness the shared and specific characteristics of different weather scenarios, we propose an \textbf{Gradient-Guided Parameter Masking} strategy. This approach involves creating three task-specific masks corresponding to rain, raindrop, and snow removal tasks. These masks enable selective modification of task-specific parameters while keeping the core model architecture frozen, thus minimizing computational demands and reducing the risk of overfitting to a single task.

\subsubsection{Creation of Task-Specific Masks}

For each weather condition \( t \in \{\text{rain}, \text{raindrop}, \text{snow}\} \), we define a mask \( M_t \) that filters parameters relevant to task \( t \). The mask \( M_t \) partitions the model's parameters \( \Theta \) into common parameters \( \Theta_c \) and task-specific parameters \( \Theta_t \):

\begin{equation}
\Theta = \Theta_c \cup \Theta_t.
\end{equation}

This partitioning allows the model to retain pre-trained general features while enabling selective adaptation to specific weather conditions.

\subsubsection{Gradient-Guided Mask Determination}

To identify the most influential parameters for each task, we perform backpropagation on the pre-trained model and calculate the gradients of all parameters with respect to the task-specific loss \( \mathcal{L}_t \). We then select the top 10\% of parameters with the highest absolute gradient values, marking their positions as true in the corresponding mask \( M_t \). Formally, for each parameter \( \theta_i \), the mask \( M_t(\theta_i) \) is defined as:

\begin{equation}
M_t(\theta_i) = 
\begin{cases} 
1, & \text{if } |\nabla_{\theta_i} \mathcal{L}_t| \geq \gamma_t \\
0, & \text{otherwise}
\end{cases}
\end{equation}

where \( \gamma_t \) is the threshold corresponding to the 90th percentile of the gradient magnitudes for task \( t \). This selection process focuses on the parameters that are most significant for the specific task, allowing efficient adaptation without modifying the frozen base model.

\subsubsection{Selective Parameter Training with Masked Parameters}

Once the masks are determined, we train only the parameters identified by the masks for each specific task \( t \), while keeping the common parameters \( \Theta_c \) frozen. The training objective for task \( t \) is defined as:

\begin{equation}
\theta_{t}^{\ast} = \operatorname{arg\,min}_{\theta} \mathcal{L}_t(f(\Theta_c, \Theta_t; X_t)),
\end{equation}

where \( \theta_{t}^{\ast} \) denotes the optimized task-specific parameters for \( t \), \( f(\cdot) \) represents the model function, and \( X_t \) is the input data for task \( t \). This selective training allows the model to adapt to each adverse weather condition effectively while maintaining computational efficiency.

\subsection{Network Architecture}

Our network architecture builds on the U-Net structure \cite{ronneberger2015u} to achieve high-quality image restoration across multiple adverse weather conditions. Unlike traditional multi-task networks \cite{zhu2023learning,li2020all}, our approach enables multi-scenario restoration within a single-task network architecture, without introducing any additional parameters.

As shown in \cref{fig:adaptive_parameter_masking}, the APMN incorporates the Gradient-Guided Parameter Masking strategy directly into the convolutional layers of the U-Net. Each convolutional layer utilizes the common parameters \( \Theta_c \) to capture general degradation patterns consistent across weather conditions, while the task-specific parameters \( \Theta_t \) are dedicated to representing unique weather-induced artifacts.

The forward pass of the network for task \( t \) is represented as:

\begin{equation}
\hat{I}_t = f_{\text{dec}}(f_{\text{enc}}(X_t; \Theta_c, \Theta_t)),
\end{equation}

where \( \hat{I}_t \) is the restored image for task \( t \), \( f_{\text{enc}} \) and \( f_{\text{dec}} \) denote the encoding and decoding functions, respectively, and \( X_t \) is the input degraded image.

\subsection{Loss Function}

To further enhance image restoration under diverse adverse weather conditions, we incorporate a composite loss function that combines pixel-wise accuracy with depth consistency. The loss function for task \( t \) is defined as:

\begin{equation}
\mathcal{L}_t = L_{1}(\hat{I}_t, Y_t) + \lambda_{\text{depth}} \| D(\hat{I}_t) - D(Y_t) \|_1,
\end{equation}

where \( L_{1}(\hat{I}_t, Y_t) \) denotes the smooth L1 loss between the restored image \( \hat{I}_t \) and the ground truth \( Y_t \), \( D(\cdot) \) represents a pre-trained depth estimation network \cite{koonce2021vgg}, and \( \lambda_{\text{depth}} \) is a weighting factor controlling the influence of the depth consistency loss. By combining pixel-wise accuracy with depth consistency, this loss formulation effectively captures both the fidelity of pixel restoration and the structural coherence of the output images.

\subsection{Inference with Gradient-Guided Parameter Mask}

During inference, the model utilizes both the frozen common parameters \( \Theta_c \) and the task-specific parameters \( \Theta_t^{\ast} \) identified by each mask \( M_t \). For a given weather condition \( t \), the model dynamically activates the corresponding task-specific parameters, allowing for efficient adaptation without requiring full retraining. The inference process is expressed as:

\begin{equation}
\hat{I}_t = f_{\text{dec}}(f_{\text{enc}}(X_t; \Theta_c, \Theta_t^{\ast})).
\end{equation}

This approach ensures that only the most relevant parameters are activated based on the input weather condition, enhancing inference efficiency and enabling real-time application.

\subsection{Summary of Approach}

Our proposed \textbf{Gradient-Guided Parameter Masking} method offers a streamlined yet effective solution for multi-scenario image restoration under adverse weather conditions. By focusing on the most impactful parameters identified through gradient-based backpropagation, our method minimizes computational overhead and reduces retraining requirements. The region-sensitive application of shared and task-specific parameters optimizes the model's configuration for robust performance across various weather scenarios, making it highly suitable for real-time applications in autonomous driving and weather-dependent image processing systems \cite{yurtsever2020survey}.

\begin{figure*}[!ht]
    \centering
    \includegraphics[width=\textwidth,clip,trim=3 3 3 3]{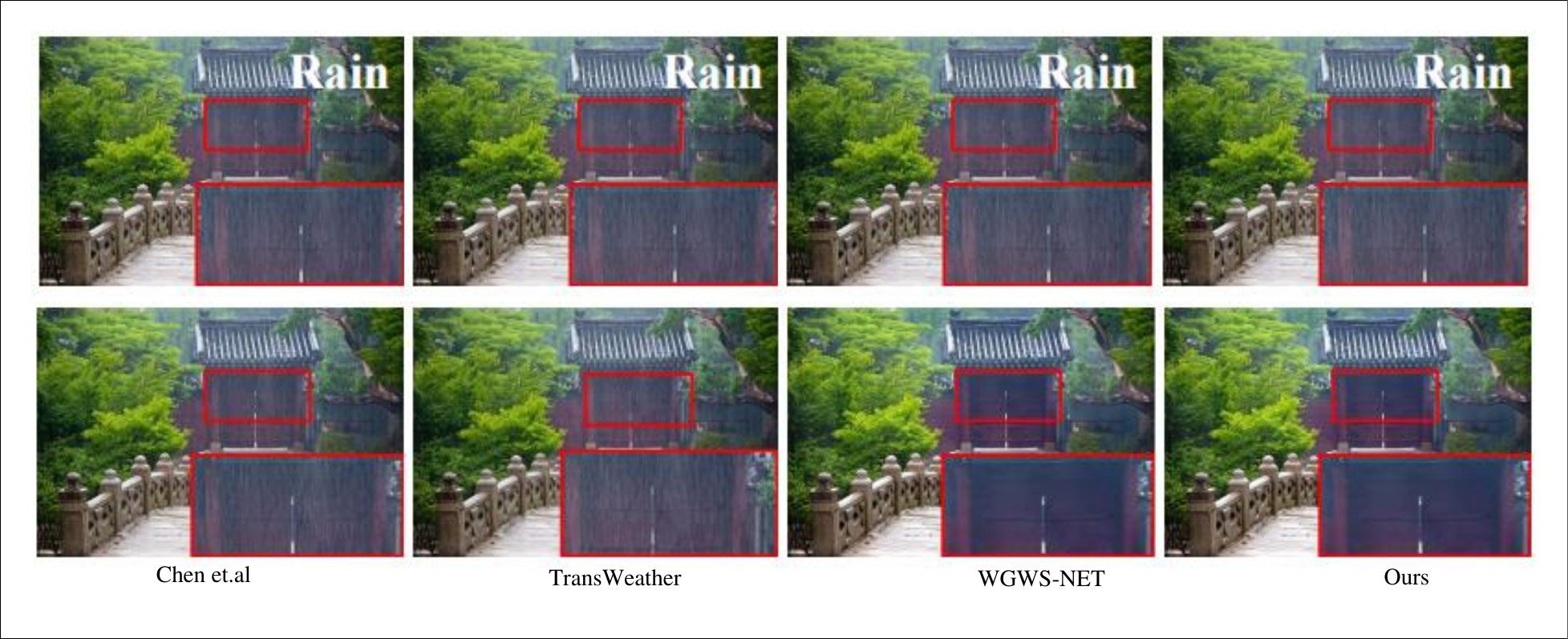} 
    \caption{Visualization comparisons with previous all-weather-removal methods under rainy weather}
    \label{visual_rain}
\end{figure*}

\section{Experiments}

We conduct extensive experiments to show the effectiveness of our proposed method. In what follows, we explain the datasets, implementation details, experimental settings, results and comparison with state-of-the-art methods.

\subsection{Datasets}
Our network is trained on a comprehensive dataset of images with diverse degradations from various adverse weather conditions, following the training set distribution of the All-in-One \cite{li2020all} Network to ensure fair comparison. The training data includes 9,000 images from the Snow100K dataset \cite{liu2018desnownet}, 1,069 images from the Raindrop dataset \cite{qian2018attentive}, and an additional 9,000 synthetic images from the Outdoor-Rain dataset \cite{hu2019depth}. Snow100K provides synthetic images simulating snow effects, while the Raindrop dataset offers real-world images affected by raindrops. The Outdoor-Rain dataset comprises synthetic images degraded by a combination of fog and rain streaks. We designate this extensive training set as "All-Weather" to emphasize its diverse weather conditions.


\subsection{Implementation Details}

We implement our method using the PyTorch framework and train it on 8 NVIDIA A100 GPUs. The training utilizes the Adam optimizer with a learning rate set to 0.0001. The network is trained for a total of 120 epochs with a batch size of 9.


\subsection{Comparison with the State-of-the-art Methods}

To comprehensively evaluate the proposed Gradient-Guided Parameter Masking framework, we conduct an extensive series of comparisons with state-of-the-art (SOTA) methods. Initially, we assess its performance against existing single restoration networks under a range of degradation scenarios, ensuring a robust evaluation across different weather conditions. Furthermore, we compare our framework with leading multi-task networks to benchmark its effectiveness in multi-domain restoration tasks. For a thorough and accurate performance analysis, we adopt Peak Signal-to-Noise Ratio (PSNR) \cite{zhang2024denoising} and Structural Similarity Index (SSIM) \cite{wang2004image} as our primary evaluation metrics. These metrics provide complementary insights, with PSNR focusing on pixel-level fidelity and SSIM assessing the perceptual quality of the restored images. This enables us to evaluate both the quantitative accuracy and the visual quality of the results, ensuring a holistic assessment of the proposed method.

\begin{table}[htbp]
    \centering
    \caption{\normalsize Comparison of Rain Removal Methods}
    \resizebox{0.5\textwidth}{!}{ 
        \begin{tabular}{c|c|c|c|c}
        \hline
        Type & Method & Venue & PSNR ↑ & SSIM ↑ \\
        \hline
        \multirow{4}{*}{Deraining}
              & pix2pix\cite{isola2017image} & CVPR'17 & 19.09 & 0.71 \\
              & HRGAN\cite{li2019heavy} & CVPR'19 & 21.56 & 0.86 \\
              & MPRNet\cite{zamir2021multi} & CVPR'21 & 21.90 & 0.85 \\
        \hline
        \multirow{5}{*}{Multi Tasks}
              & All-in-One\cite{li2020all} & CVPR'20 & 24.71 & 0.90 \\
              & TransWeather\cite{valanarasu2022transweather} & CVPR'22 & 23.18 & 0.84 \\
              & Chen et al.\cite{chen2022learning} & CVPR'22 & 23.94 & 0.85 \\
              & Ours & - & \textbf{29.22} & \textbf{0.91} \\
        \hline
    \end{tabular}
    }
    \label{tab:Deraining}
\end{table}

\begin{table}[htbp]
    \centering
    \caption{\normalsize Comparison of RainDrop Removal Methods}
    \resizebox{0.5\textwidth}{!}{ 
        \begin{tabular}{c|c|c|c|c}
        \hline
        Type & Method & Venue & PSNR ↑ & SSIM ↑ \\
        \hline
        \multirow{4}{*}{RainDrop} 
              & Pix2pix\cite{isola2017image} & CVPR'17 & 28.02 & 0.85 \\
              & Attn.GAN\cite{qian2018attentive} & CVPR'18 & 30.55 & 0.90 \\
              & Quan et al.\cite{quan2019deep} & ICCV'19 & 31.44 & 0.93 \\
              & CCN\cite{quan2021removing} & CVPR'21 & 31.34 & 0.95 \\
        \hline
        \multirow{5}{*}{Multi Tasks} 
              & All-in-One\cite{li2020all} & CVPR'20 & \textbf{31.12} & \textbf{0.93} \\
              & TransWeather\cite{valanarasu2022transweather} & CVPR'22 & 28.98 & 0.90 \\
              & Chen et al.\cite{chen2022learning} & CVPR'22 & 30.75 & 0.91 \\
              & Ours & - & 30.76 & 0.91 \\
        \hline
    \end{tabular}
    }
    \label{tab:raindrop_removal}
\end{table}

\begin{table}[htbp]
    \centering
    \caption{\normalsize Comparison of Snow Removal Methods}
    \resizebox{0.5\textwidth}{!}{ 
        \begin{tabular}{c|c|c|c|c}
        \hline
        Type & Method & Venue & PSNR ↑ & SSIM ↑ \\
        \hline
        \multirow{4}{*}{Desnowing} 
              & DetailsNet \cite{fu2017removing} & CVPR '17 & 19.18 & 0.75 \\ 
              & DesnowNet \cite{liu2018desnownet} & TIP '18 & 27.17 & 0.90 \\ 
              & JSTASR \cite{chen2020jstasr} & ECCV '20 & 25.32 & 0.81 \\ 
              & DDMSNET \cite{zhang2021deep} & TIP '21 & 28.85 & 0.88 \\ \hline
        \hline
        \multirow{5}{*}{Multi Tasks}
              & All-in-One \cite{li2020all} & CVPR '20 & 28.33 & 0.88 \\ 
              & TransWeather \cite{valanarasu2022transweather} & CVPR '22 & 27.80 & 0.85 \\ 
              & Chen et al.\cite{chen2022learning}  & CVPR '22 & 29.27 & 0.88 \\ 
              & Ours & - & \textbf{29.56} & \textbf{0.89} \\ \hline
              
        \hline
    \end{tabular}
    }
    \label{tab:snow}
\end{table}


\begin{figure*}[!t]
    \centering
    \includegraphics[width=\textwidth,clip,trim=3 3 3 3]{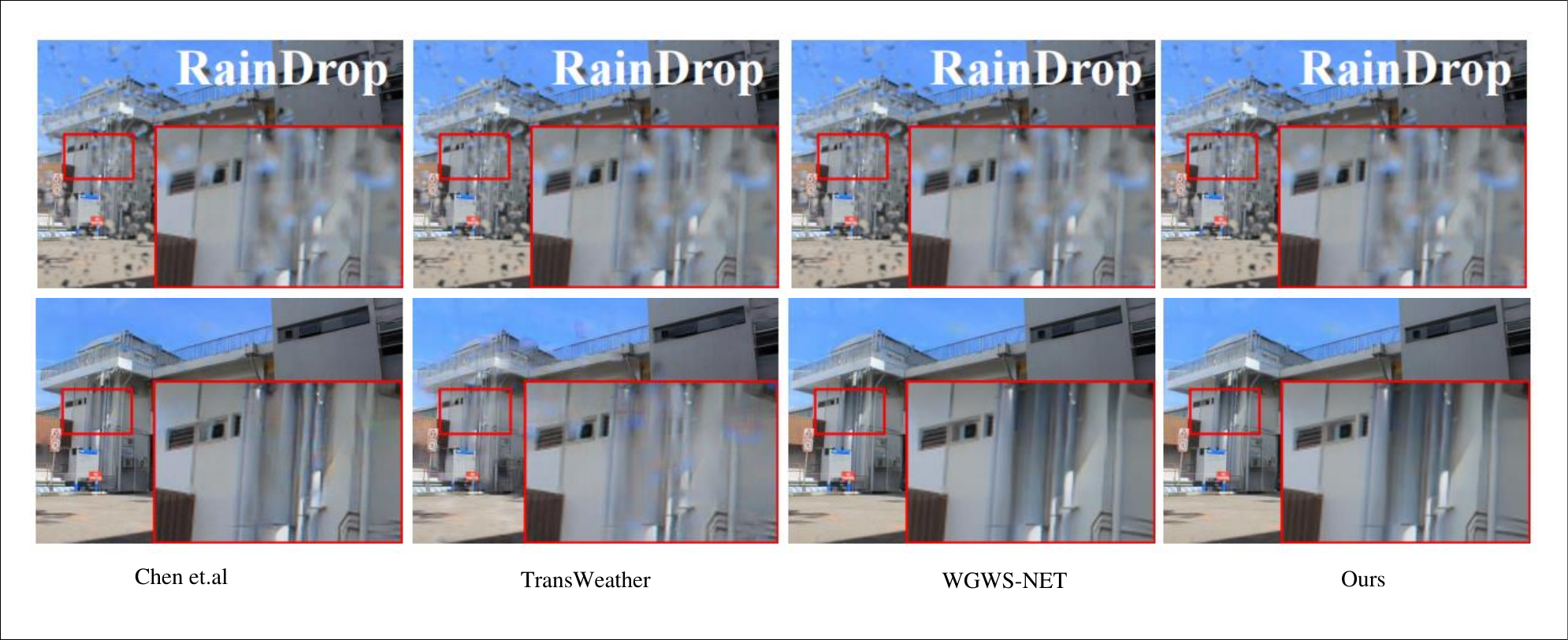} 
    \caption{Visualization comparisons with previous all-weather-removal methods under raindrop weather}
    \label{visual_raindrop}
\end{figure*}

\begin{figure*}[!ht]
    \centering
    \includegraphics[width=\textwidth,clip,trim=3 3 3 3]{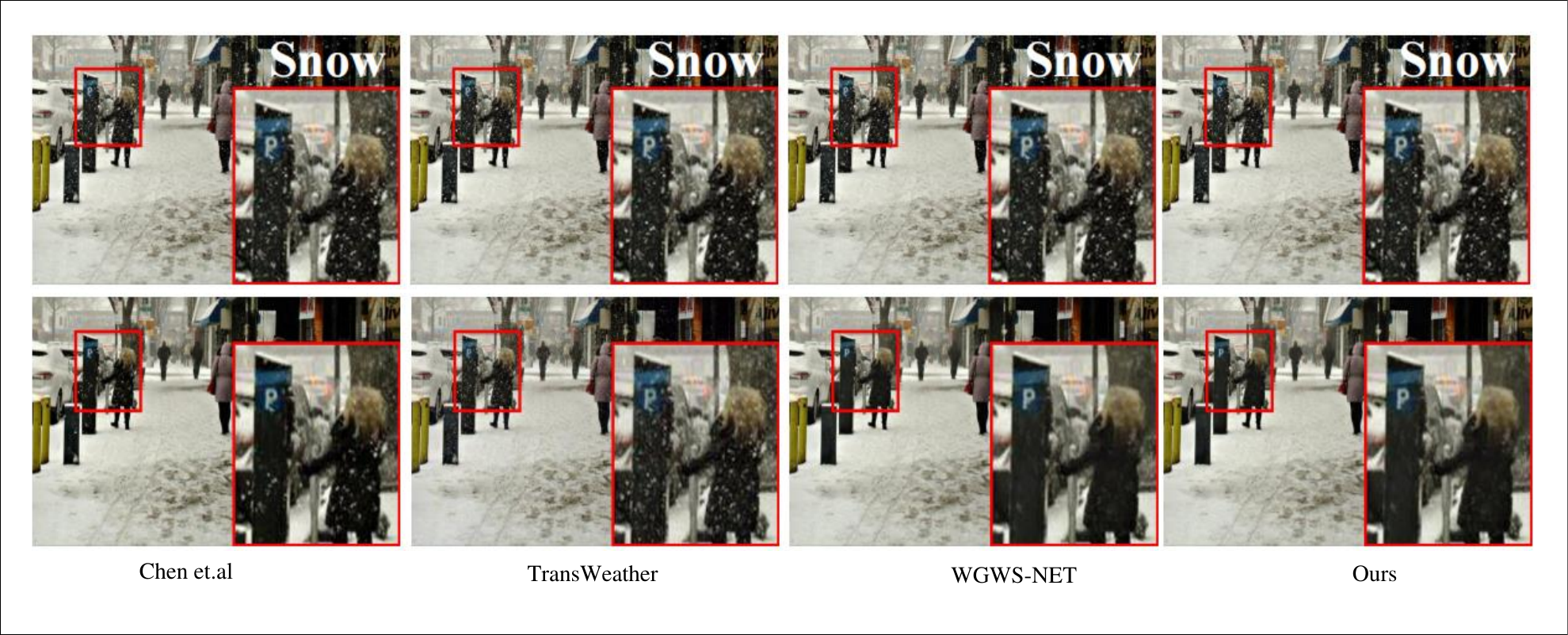} 
    \caption{Visualization comparisons with previous all-weather-removal methods under snowy weather}
    \label{visual_snow}
\end{figure*}

As shown in \cref{tab:Deraining},\cref{tab:raindrop_removal} and \cref{tab:snow}, our proposed method achieves strong performance across all tasks. When compared to existing task-specific models and multi-task networks, our approach outperforms all models except \cite{zhu2023learning}. This is primarily due to the introduction of the Gradient-Guided Mask strategy, which allows the model's parameters to be adaptively selected for more precise optimization and updating. This strategy helps mitigate the impact of gradient conflicts across different tasks, improving overall performance. Moreover, our method partitions the mask based on gradients to selectively optimize and update parameters, ensuring that more important parameters are updated with greater focus. However, as demonstrated in \cref{tab:para}, our method requires significantly fewer parameters than other approaches, which leads to a slight reduction in the model’s representational capacity. This reduction in capacity results in slightly lower performance compared to \cite{zhu2023learning}.

Moreover, we provide visual comparisons on the various kinds of weather in \cref{visual_rain}, \cref{visual_raindrop}, \cref{visual_snow}. It is evident that our results successfully preserve background details and remove multiple weather artifacts.

\section{Conclusion}
In this paper, we proposed the Gradient-Guided Parameter Mask (APM), a novel approach to multi-scenario image restoration under adverse weather conditions. By leveraging task-specific masks, our method enables the adaptive selection and fine-tuned optimization of parameters without increasing the overall model complexity. Through extensive experimentation, we demonstrated that our framework outperforms existing state-of-the-art methods across a variety of weather scenarios, including rain, snow, and raindrops, while maintaining computational efficiency suitable for real-time applications. Our proposed method effectively mitigates gradient conflicts between tasks by selectively optimizing only the most relevant parameters for each specific weather condition. This approach not only reduces model size but also enhances inference speed, making it highly applicable to real-time systems such as autonomous vehicles and weather-dependent image processing. 

\section{Discussion}

\begin{figure}[!t]
    \centering
    \includegraphics[width=\linewidth,clip,trim=3 3 3 3]{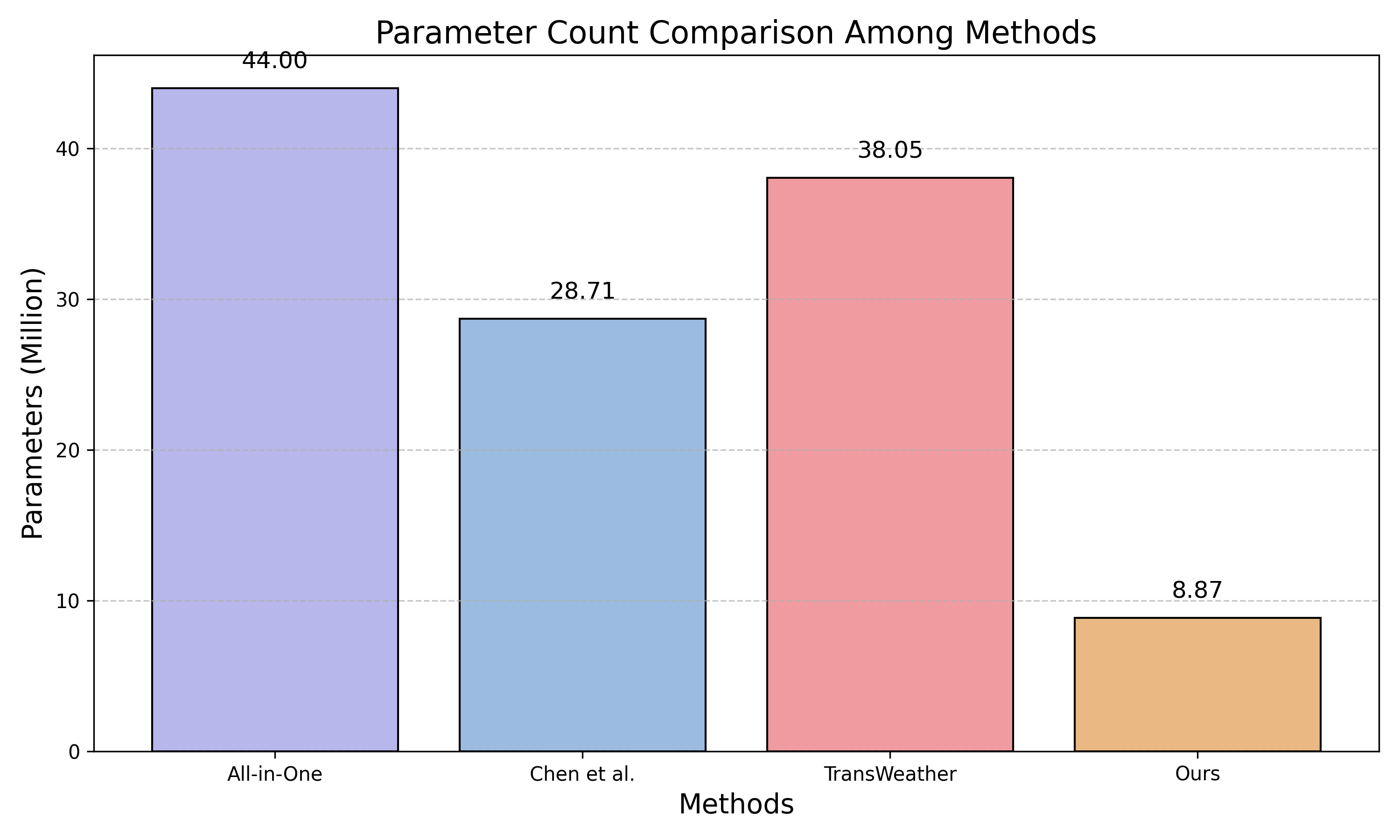} 
    \caption{Comparing the number of parameters of different models}
    \label{tab:para}
\end{figure}

Our proposed method demonstrates several key advantages over existing approaches. Firstly, due to its design that avoids adding extra parameters, our method maintains a significantly smaller parameter size compared to other methods, as illustrated in \cref{tab:para}. This reduction in parameter scale is especially valuable for applications in fields such as autonomous driving, where computational resources on end devices are often limited. A smaller model size reduces both memory and computational load, making our method highly suitable for deployment in real-time, resource-constrained environments.

Secondly, our method exhibits strong generalization capabilities. Since we do not alter the underlying model architecture or parameters, our approach is compatible with a wide range of state-of-the-art (SOTA) models. This plug-and-play nature enables seamless integration with existing models without the need for extensive modification, enhancing the versatility and ease of use of our method in diverse applications. This adaptability, combined with computational efficiency, positions our approach as a practical and scalable solution for multi-scenario adverse weather restoration in various real-world applications.

{
    \small
    \bibliographystyle{unsrt}
    \bibliography{main}

\begin{thebibliography}{10}

\bibitem{li2020all}
Ruoteng Li, Robby~T Tan, and Loong-Fah Cheong.
\newblock All in one bad weather removal using architectural search.
\newblock In {\em Proceedings of the IEEE/CVF conference on computer vision and pattern recognition}, pages 3175--3185, 2020.

\bibitem{zhu2023learning}
Yurui Zhu, Tianyu Wang, Xueyang Fu, Xuanyu Yang, Xin Guo, Jifeng Dai, Yu~Qiao, and Xiaowei Hu.
\newblock Learning weather-general and weather-specific features for image restoration under multiple adverse weather conditions.
\newblock In {\em Proceedings of the IEEE/CVF conference on computer vision and pattern recognition}, pages 21747--21758, 2023.

\bibitem{zhang2023perception}
Yuxiao Zhang, Alexander Carballo, Hanting Yang, and Kazuya Takeda.
\newblock Perception and sensing for autonomous vehicles under adverse weather conditions: A survey.
\newblock {\em ISPRS Journal of Photogrammetry and Remote Sensing}, 196:146--177, 2023.

\bibitem{vargas2021overview}
Jorge Vargas, Suleiman Alsweiss, Onur Toker, Rahul Razdan, and Joshua Santos.
\newblock An overview of autonomous vehicles sensors and their vulnerability to weather conditions.
\newblock {\em Sensors}, 21(16):5397, 2021.

\bibitem{rothmeier2021let}
Thomas Rothmeier and Werner Huber.
\newblock Let it snow: On the synthesis of adverse weather image data.
\newblock In {\em 2021 IEEE International Intelligent Transportation Systems Conference (ITSC)}, pages 3300--3306. IEEE, 2021.

\bibitem{meister2022designing}
Philippe Meister, Jack Miller, Kexin Wang, Michael~C Dorneich, Eliot Winer, Lori~J Brown, and Geoffrey Whitehurst.
\newblock Designing three-dimensional augmented reality weather visualizations to enhance general aviation weather education.
\newblock {\em IEEE Transactions on Professional Communication}, 65(2):321--336, 2022.

\bibitem{zeng2020hyperspectral}
Haijin Zeng, Xiaozhen Xie, Haojie Cui, Yuan Zhao, and Jifeng Ning.
\newblock Hyperspectral image restoration via cnn denoiser prior regularized low-rank tensor recovery.
\newblock {\em Computer Vision and Image Understanding}, 197:103004, 2020.

\bibitem{soh2022variational}
Jae~Woong Soh and Nam~Ik Cho.
\newblock Variational deep image restoration.
\newblock {\em IEEE Transactions on Image Processing}, 31:4363--4376, 2022.

\bibitem{su2022survey}
Jingwen Su, Boyan Xu, and Hujun Yin.
\newblock A survey of deep learning approaches to image restoration.
\newblock {\em Neurocomputing}, 487:46--65, 2022.

\bibitem{ali2023vision}
Anas~M Ali, Bilel Benjdira, Anis Koubaa, Walid El-Shafai, Zahid Khan, and Wadii Boulila.
\newblock Vision transformers in image restoration: A survey.
\newblock {\em Sensors}, 23(5):2385, 2023.

\bibitem{zamir2022restormer}
Syed~Waqas Zamir, Aditya Arora, Salman Khan, Munawar Hayat, Fahad~Shahbaz Khan, and Ming-Hsuan Yang.
\newblock Restormer: Efficient transformer for high-resolution image restoration.
\newblock In {\em Proceedings of the IEEE/CVF conference on computer vision and pattern recognition}, pages 5728--5739, 2022.

\bibitem{liang2021swinir}
Jingyun Liang, Jiezhang Cao, Guolei Sun, Kai Zhang, Luc Van~Gool, and Radu Timofte.
\newblock Swinir: Image restoration using swin transformer.
\newblock In {\em Proceedings of the IEEE/CVF international conference on computer vision}, pages 1833--1844, 2021.

\bibitem{karavarsamis2022survey}
Sotiris Karavarsamis, Ioanna Gkika, Vasileios Gkitsas, Konstantinos Konstantoudakis, and Dimitrios Zarpalas.
\newblock A survey of deep learning-based image restoration methods for enhancing situational awareness at disaster sites: the cases of rain, snow and haze.
\newblock {\em Sensors}, 22(13):4707, 2022.

\bibitem{rutten2023deep}
Linda Rutten.
\newblock Deep learning for weather condition adaptation in autonomous vehicles.
\newblock {\em Journal of Artificial Intelligence Research and Applications}, 3(1):274--306, 2023.

\bibitem{shi2023spatial}
Dapai Shi, Jingyuan Zhao, Zhenghong Wang, Heng Zhao, Junbin Wang, Yubo Lian, and Andrew~F Burke.
\newblock Spatial-temporal self-attention transformer networks for battery state of charge estimation.
\newblock {\em Electronics}, 12(12):2598, 2023.

\bibitem{choi2023transformer}
Sanghyuk~Roy Choi and Minhyeok Lee.
\newblock Transformer architecture and attention mechanisms in genome data analysis: a comprehensive review.
\newblock {\em Biology}, 12(7):1033, 2023.

\bibitem{yurtsever2020survey}
Ekim Yurtsever, Jacob Lambert, Alexander Carballo, and Kazuya Takeda.
\newblock A survey of autonomous driving: Common practices and emerging technologies.
\newblock {\em IEEE access}, 8:58443--58469, 2020.

\bibitem{kozar2023recovery}
Daniel Kozar, Xiaoli Dong, and Li~Li.
\newblock The recovery of river chemistry from acid rain in the mississippi river basin amid intensifying anthropogenic activities and climate change.
\newblock {\em Science of the Total Environment}, 897:165311, 2023.

\bibitem{yang2022rain}
Fei Yang, Jianfeng Ren, Zheng Lu, Jialu Zhang, and Qian Zhang.
\newblock Rain-component-aware capsule-gan for single image de-raining.
\newblock {\em Pattern Recognition}, 123:108377, 2022.

\bibitem{chen2022lightweightderain}
Zheng Chen, Xiaojun Bi, Yu~Zhang, Jianyu Yue, and Haibo Wang.
\newblock Lightweightderain: learning a lightweight multi-scale high-order feedback network for single image de-raining.
\newblock {\em Neural Computing and Applications}, pages 1--18, 2022.

\bibitem{fu2023continual}
Xueyang Fu, Jie Xiao, Yurui Zhu, Aiping Liu, Feng Wu, and Zheng-Jun Zha.
\newblock Continual image deraining with hypergraph convolutional networks.
\newblock {\em IEEE Transactions on Pattern Analysis and Machine Intelligence}, 45(8):9534--9551, 2023.

\bibitem{hu2021single}
Xiaowei Hu, Lei Zhu, Tianyu Wang, Chi-Wing Fu, and Pheng-Ann Heng.
\newblock Single-image real-time rain removal based on depth-guided non-local features.
\newblock {\em IEEE Transactions on Image Processing}, 30:1759--1770, 2021.

\bibitem{jiang2020multi}
Kui Jiang, Zhongyuan Wang, Peng Yi, Chen Chen, Baojin Huang, Yimin Luo, Jiayi Ma, and Junjun Jiang.
\newblock Multi-scale progressive fusion network for single image deraining.
\newblock In {\em Proceedings of the IEEE/CVF conference on computer vision and pattern recognition}, pages 8346--8355, 2020.

\bibitem{wang2020model}
Hong Wang, Qi~Xie, Qian Zhao, and Deyu Meng.
\newblock A model-driven deep neural network for single image rain removal.
\newblock In {\em Proceedings of the IEEE/CVF conference on computer vision and pattern recognition}, pages 3103--3112, 2020.

\bibitem{xiao2022image}
Jie Xiao, Xueyang Fu, Aiping Liu, Feng Wu, and Zheng-Jun Zha.
\newblock Image de-raining transformer.
\newblock {\em IEEE Transactions on Pattern Analysis and Machine Intelligence}, 45(11):12978--12995, 2022.

\bibitem{xiao2021improving}
Jie Xiao, Man Zhou, Xueyang Fu, Aiping Liu, and Zheng-Jun Zha.
\newblock Improving de-raining generalization via neural reorganization.
\newblock In {\em Proceedings of the IEEE/CVF International Conference on Computer Vision}, pages 4987--4996, 2021.

\bibitem{yang2020single}
Wenhan Yang, Robby~T Tan, Shiqi Wang, Yuming Fang, and Jiaying Liu.
\newblock Single image deraining: From model-based to data-driven and beyond.
\newblock {\em IEEE Transactions on pattern analysis and machine intelligence}, 43(11):4059--4077, 2020.

\bibitem{chen2020jstasr}
Wei-Ting Chen, Hao-Yu Fang, Jian-Jiun Ding, Cheng-Che Tsai, and Sy-Yen Kuo.
\newblock Jstasr: Joint size and transparency-aware snow removal algorithm based on modified partial convolution and veiling effect removal.
\newblock In {\em Computer Vision--ECCV 2020: 16th European Conference, Glasgow, UK, August 23--28, 2020, Proceedings, Part XXI 16}, pages 754--770. Springer, 2020.

\bibitem{zhang2021deep}
Kaihao Zhang, Rongqing Li, Yanjiang Yu, Wenhan Luo, and Changsheng Li.
\newblock Deep dense multi-scale network for snow removal using semantic and depth priors.
\newblock {\em IEEE Transactions on Image Processing}, 30:7419--7431, 2021.

\bibitem{li2021online}
Minghan Li, Xiangyong Cao, Qian Zhao, Lei Zhang, and Deyu Meng.
\newblock Online rain/snow removal from surveillance videos.
\newblock {\em IEEE Transactions on Image Processing}, 30:2029--2044, 2021.

\bibitem{jaw2020desnowgan}
Da-Wei Jaw, Shih-Chia Huang, and Sy-Yen Kuo.
\newblock Desnowgan: An efficient single image snow removal framework using cross-resolution lateral connection and gans.
\newblock {\em IEEE Transactions on Circuits and Systems for Video Technology}, 31(4):1342--1350, 2020.

\bibitem{zhang2021dual}
Kaihao Zhang, Dongxu Li, Wenhan Luo, and Wenqi Ren.
\newblock Dual attention-in-attention model for joint rain streak and raindrop removal.
\newblock {\em IEEE Transactions on Image Processing}, 30:7608--7619, 2021.

\bibitem{luo2020weakly}
Wenjie Luo, Jianhuang Lai, and Xiaohua Xie.
\newblock Weakly supervised learning for raindrop removal on a single image.
\newblock {\em IEEE Transactions on Circuits and Systems for Video Technology}, 31(5):1673--1683, 2020.

\bibitem{yan2022feature}
Wending Yan, Lu~Xu, Wenhan Yang, and Robby~T Tan.
\newblock Feature-aligned video raindrop removal with temporal constraints.
\newblock {\em IEEE Transactions on Image Processing}, 31:3440--3448, 2022.

\bibitem{lin2020x}
Jiamin Lin and Longquan Dai.
\newblock X-net for single image raindrop removal.
\newblock In {\em 2020 IEEE International Conference on Image Processing (ICIP)}, pages 1003--1007. IEEE, 2020.

\bibitem{ragini2022rain}
Thatikonda Ragini and Kodali Prakash.
\newblock Rain streak removal via spatio-channel based spectral graph cnn for image deraining.
\newblock In {\em International Conference on Computer Vision and Image Processing}, pages 149--160. Springer, 2022.

\bibitem{quan2021removing}
Ruijie Quan, Xin Yu, Yuanzhi Liang, and Yi~Yang.
\newblock Removing raindrops and rain streaks in one go.
\newblock In {\em Proceedings of the IEEE/CVF conference on computer vision and pattern recognition}, pages 9147--9156, 2021.

\bibitem{ding2021rain}
Yuyang Ding, Mingyue Li, Tao Yan, Fan Zhang, Yuan Liu, and Rynson~WH Lau.
\newblock Rain streak removal from light field images.
\newblock {\em IEEE Transactions on Circuits and Systems for Video Technology}, 32(2):467--482, 2021.

\bibitem{zhang2019image}
He~Zhang, Vishwanath Sindagi, and Vishal~M Patel.
\newblock Image de-raining using a conditional generative adversarial network.
\newblock {\em IEEE transactions on circuits and systems for video technology}, 30(11):3943--3956, 2019.

\bibitem{li2019heavy}
Ruoteng Li, Loong-Fah Cheong, and Robby~T Tan.
\newblock Heavy rain image restoration: Integrating physics model and conditional adversarial learning.
\newblock In {\em Proceedings of the IEEE/CVF conference on computer vision and pattern recognition}, pages 1633--1642, 2019.

\bibitem{huang2021memory}
Huaibo Huang, Aijing Yu, and Ran He.
\newblock Memory oriented transfer learning for semi-supervised image deraining.
\newblock In {\em Proceedings of the IEEE/CVF conference on computer vision and pattern recognition}, pages 7732--7741, 2021.

\bibitem{ye2022unsupervised}
Yuntong Ye, Changfeng Yu, Yi~Chang, Lin Zhu, Xi-Le Zhao, Luxin Yan, and Yonghong Tian.
\newblock Unsupervised deraining: Where contrastive learning meets self-similarity.
\newblock In {\em Proceedings of the IEEE/CVF conference on computer vision and pattern recognition}, pages 5821--5830, 2022.

\bibitem{huang2022deep}
Jie Huang, Yajing Liu, Feng Zhao, Keyu Yan, Jinghao Zhang, Yukun Huang, Man Zhou, and Zhiwei Xiong.
\newblock Deep fourier-based exposure correction network with spatial-frequency interaction.
\newblock In {\em European Conference on Computer Vision}, pages 163--180. Springer, 2022.

\bibitem{ni2021controlling}
Siqi Ni, Xueyun Cao, Tao Yue, and Xuemei Hu.
\newblock Controlling the rain: From removal to rendering.
\newblock In {\em Proceedings of the IEEE/CVF Conference on Computer Vision and Pattern Recognition}, pages 6328--6337, 2021.

\bibitem{lin2020sequential}
Chih-Yang Lin, Zhuang Tao, Ai-Sheng Xu, Li-Wei Kang, and Fityanul Akhyar.
\newblock Sequential dual attention network for rain streak removal in a single image.
\newblock {\em IEEE Transactions on Image Processing}, 29:9250--9265, 2020.

\bibitem{tang2022multi}
Lai~Meng Tang, Li~Hong~Idris Lim, and Paul Siebert.
\newblock Multi-scale rain removal across multiple frequencies sub-bands using ms-cyclegans.
\newblock In {\em International Symposium on Automation, Mechanical and Design Engineering}, pages 233--245. Springer, 2022.

\bibitem{9040628}
Tao Yang, Xiaofei Chang, Hang Su, Nathan Crombez, Yassine Ruichek, Tomas Krajnik, and Zhi Yan.
\newblock Raindrop removal with light field image using image inpainting.
\newblock {\em IEEE Access}, 8:58416--58426, 2020.

\bibitem{8578361}
Rui Qian, Robby~T. Tan, Wenhan Yang, Jiajun Su, and Jiaying Liu.
\newblock Attentive generative adversarial network for raindrop removal from a single image.
\newblock In {\em 2018 IEEE/CVF Conference on Computer Vision and Pattern Recognition}, pages 2482--2491, 2018.

\bibitem{9577896}
Ruijie Quan, Xin Yu, Yuanzhi Liang, and Yi~Yang.
\newblock Removing raindrops and rain streaks in one go.
\newblock In {\em 2021 IEEE/CVF Conference on Computer Vision and Pattern Recognition (CVPR)}, pages 9143--9152, 2021.

\bibitem{10378538}
Sixiang Chen, Tian Ye, Jinbin Bai, Erkang Chen, Jun Shi, and Lei Zhu.
\newblock Sparse sampling transformer with uncertainty-driven ranking for unified removal of raindrops and rain streaks.
\newblock In {\em 2023 IEEE/CVF International Conference on Computer Vision (ICCV)}, pages 13060--13071, 2023.

\bibitem{9707527}
Xu~Yan and Yuan~Ren Loke.
\newblock Raingan: Unsupervised raindrop removal via decomposition and composition.
\newblock In {\em 2022 IEEE/CVF Winter Conference on Applications of Computer Vision Workshops (WACVW)}, pages 14--23, 2022.

\bibitem{9897458}
Yan Shen, Jiange Xu, Xiaotao Shao, Jinbiao Zhu, and Xinmin Wang.
\newblock Toward snow removal via the diversity and complexity of snow image.
\newblock In {\em 2022 IEEE International Conference on Image Processing (ICIP)}, pages 1966--1970, 2022.

\bibitem{10070780}
Guisik Kim, Sungmin Cho, Dokyeong Kwon, Seo~Hyeon Lee, and Junseok Kwon.
\newblock Dual gradient based snow attentive desnowing.
\newblock {\em IEEE Access}, 11:26086--26098, 2023.

\bibitem{karavarsamis2022cross}
Sotiris Karavarsamis, Alexandros Doumanoglou, Konstantinos Konstantoudakis, and Dimitrios Zarpalas.
\newblock Cross-stitched multi-task dual recursive networks for unified single image deraining and desnowing.
\newblock In {\em 2022 IEEE 8th World Forum on Internet of Things (WF-IoT)}, pages 1--6. IEEE, 2022.

\bibitem{sun2022landslide}
Shu Sun, Xiaoping Wang, Junnan Li, and Cheng Lian.
\newblock Landslide evolution state prediction and down-level control based on multi-task learning.
\newblock {\em Knowledge-Based Systems}, 238:107884, 2022.

\bibitem{10196308}
Tao Gao, Yuanbo Wen, Kaihao Zhang, Jing Zhang, Ting Chen, Lidong Liu, and Wenhan Luo.
\newblock Frequency-oriented efficient transformer for all-in-one weather-degraded image restoration.
\newblock {\em IEEE Transactions on Circuits and Systems for Video Technology}, 34(3):1886--1899, 2024.

\bibitem{10506517}
Shugo Yamashita and Masaaki Ikehara.
\newblock Multiple adverse weather removal using masked-based pre-training and dual-pooling adaptive convolution.
\newblock {\em IEEE Access}, 12:58057--58066, 2024.

\bibitem{Wang2024LRBT}
Pengyu Wang, Hongqing Zhu, Huaqi Zhang, and Suyi Yang.
\newblock Lrb-t: local reasoning back-projection transformer for the removal of bad weather effects in images.
\newblock {\em Neural Computing and Applications}, 36(3):773--789, 2024.

\bibitem{narasimhan2003contrast}
Srinivasa~G. Narasimhan and Shree~K. Nayar.
\newblock Contrast restoration of weather degraded images.
\newblock {\em IEEE transactions on pattern analysis and machine intelligence}, 25(6):713--724, 2003.

\bibitem{an2024comprehensive}
Shunmin An, Xixia Huang, Lujia Cao, and Linling Wang.
\newblock A comprehensive survey on image dehazing for different atmospheric scattering models.
\newblock {\em Multimedia Tools and Applications}, 83(14):40963--40993, 2024.

\bibitem{nayar1999vision}
Shree~K Nayar and Srinivasa~G Narasimhan.
\newblock Vision in bad weather.
\newblock In {\em Proceedings of the seventh IEEE international conference on computer vision}, volume~2, pages 820--827. IEEE, 1999.

\bibitem{ronneberger2015u}
Olaf Ronneberger, Philipp Fischer, and Thomas Brox.
\newblock U-net: Convolutional networks for biomedical image segmentation.
\newblock In {\em Medical image computing and computer-assisted intervention--MICCAI 2015: 18th international conference, Munich, Germany, October 5-9, 2015, proceedings, part III 18}, pages 234--241. Springer, 2015.

\bibitem{koonce2021vgg}
Brett Koonce and Brett Koonce.
\newblock Vgg network.
\newblock {\em Convolutional Neural Networks with Swift for Tensorflow: Image Recognition and Dataset Categorization}, pages 35--50, 2021.

\bibitem{liu2018desnownet}
Yun-Fu Liu, Da-Wei Jaw, Shih-Chia Huang, and Jenq-Neng Hwang.
\newblock Desnownet: Context-aware deep network for snow removal.
\newblock {\em IEEE Transactions on Image Processing}, 27(6):3064--3073, 2018.

\bibitem{qian2018attentive}
Rui Qian, Robby~T Tan, Wenhan Yang, Jiajun Su, and Jiaying Liu.
\newblock Attentive generative adversarial network for raindrop removal from a single image.
\newblock In {\em Proceedings of the IEEE conference on computer vision and pattern recognition}, pages 2482--2491, 2018.

\bibitem{hu2019depth}
Xiaowei Hu, Chi-Wing Fu, Lei Zhu, and Pheng-Ann Heng.
\newblock Depth-attentional features for single-image rain removal.
\newblock In {\em Proceedings of the IEEE/CVF Conference on computer vision and pattern recognition}, pages 8022--8031, 2019.

\bibitem{zhang2024denoising}
Boyan Zhang, Yingqi Zhang, Binjie Wang, Xin He, Fan Zhang, and Xinhong Zhang.
\newblock Denoising swin transformer and perceptual peak signal-to-noise ratio for low-dose ct image denoising.
\newblock {\em Measurement}, 227:114303, 2024.

\bibitem{wang2004image}
Zhou Wang, Alan~C Bovik, Hamid~R Sheikh, and Eero~P Simoncelli.
\newblock Image quality assessment: from error visibility to structural similarity.
\newblock {\em IEEE transactions on image processing}, 13(4):600--612, 2004.

\bibitem{isola2017image}
Phillip Isola, Jun-Yan Zhu, Tinghui Zhou, and Alexei~A Efros.
\newblock Image-to-image translation with conditional adversarial networks.
\newblock In {\em Proceedings of the IEEE conference on computer vision and pattern recognition}, pages 1125--1134, 2017.

\bibitem{zamir2021multi}
Syed~Waqas Zamir, Aditya Arora, Salman Khan, Munawar Hayat, Fahad~Shahbaz Khan, Ming-Hsuan Yang, and Ling Shao.
\newblock Multi-stage progressive image restoration.
\newblock In {\em Proceedings of the IEEE/CVF conference on computer vision and pattern recognition}, pages 14821--14831, 2021.

\bibitem{valanarasu2022transweather}
Jeya Maria~Jose Valanarasu, Rajeev Yasarla, and Vishal~M Patel.
\newblock Transweather: Transformer-based restoration of images degraded by adverse weather conditions.
\newblock In {\em Proceedings of the IEEE/CVF Conference on Computer Vision and Pattern Recognition}, pages 2353--2363, 2022.

\bibitem{chen2022learning}
Wei-Ting Chen, Zhi-Kai Huang, Cheng-Che Tsai, Hao-Hsiang Yang, Jian-Jiun Ding, and Sy-Yen Kuo.
\newblock Learning multiple adverse weather removal via two-stage knowledge learning and multi-contrastive regularization: Toward a unified model.
\newblock In {\em Proceedings of the IEEE/CVF Conference on Computer Vision and Pattern Recognition}, pages 17653--17662, 2022.

\bibitem{quan2019deep}
Yuhui Quan, Shijie Deng, Yixin Chen, and Hui Ji.
\newblock Deep learning for seeing through window with raindrops.
\newblock In {\em Proceedings of the IEEE/CVF International Conference on Computer Vision}, pages 2463--2471, 2019.

\bibitem{fu2017removing}
Xueyang Fu, Jiabin Huang, Delu Zeng, Yue Huang, Xinghao Ding, and John Paisley.
\newblock Removing rain from single images via a deep detail network.
\newblock In {\em Proceedings of the IEEE conference on computer vision and pattern recognition}, pages 3855--3863, 2017.

\end{thebibliography}
}


\end{document}